\documentclass[letterpaper, 10 pt, conference]{ieeeconf}  % Comment this line out if you need a4paper

\IEEEoverridecommandlockouts                              % This command is only needed if 
\makeatletter
\let\NAT@parse\undefined
\makeatother
\usepackage{graphics} % for pdf, bitmapped graphics files
\usepackage{graphicx}
\usepackage{grffile} %file name with multi-dots
\usepackage{amsmath} % assumes amsmath package installed
\usepackage{amssymb}  % assumes amsmath package installed
\usepackage{color}
\usepackage{mathtools} 
\usepackage{booktabs,ragged2e}
\usepackage{breqn}
\usepackage{textgreek}
\usepackage{arydshln}
\usepackage[flushleft]{threeparttable}
\usepackage{tabularx,booktabs}
\usepackage{graphics}
\usepackage{amsmath}
\usepackage{gensymb}
\usepackage[normalem]{ulem} % for \sout
\usepackage{pifont}
\usepackage{booktabs}
\usepackage{multirow}

\usepackage{lineno}
\usepackage{cite}
\usepackage{bm}
\usepackage{url}

\usepackage{makecell}
\usepackage[square, comma, numbers,sort&compress]{natbib}
\usepackage{multirow}
\let\labelindent\relax
\usepackage{enumitem}
\usepackage[table,dvipsnames]{xcolor}
\usepackage[pagebackref=true,breaklinks=true,colorlinks,bookmarks=false]{hyperref}
\usepackage{comment}
\usepackage{svg}
\hypersetup{
linkcolor=BrickRed
,citecolor=Green
,filecolor=Mulberry
,urlcolor=NavyBlue
,menucolor=BrickRed
,runcolor=Mulberry
,linkbordercolor=BrickRed
,citebordercolor=Green
,filebordercolor=Mulberry
,urlbordercolor=NavyBlue
,menubordercolor=BrickRed
,runbordercolor=Mulberry
}

\usepackage[colorinlistoftodos]{todonotes}
\usepackage[normalem]{ulem} 
\newcommand{\uksang}[1]{\todo[inline, color=blue!20]{Uksang: #1}}
\renewcommand{\uksang}[1]{}

\renewcommand{\todo}[1]{{{\textbf{{\color{Red}#1}}}}}

\usepackage{balance}
\usepackage{xcolor}
\usepackage{makecell}% 
\usepackage{natbib}
\usepackage{ulem}
\usepackage{algpseudocode}
\usepackage{algorithm}
\usepackage{hyperref}

\title{\LARGE \bf
RoPotter: Toward Robotic Pottery and Deformable Object Manipulation with Structural Priors
}
\author{\small Uksang Yoo \textsuperscript{1\ding{41}*}% <-this % stops a space
, Adam Hung\textsuperscript{2*}, Jonathan Francis\textsuperscript{1,3}, Jean Oh\textsuperscript{1}, and Jeffrey Ichnowski\textsuperscript{1}
\thanks{$^{*}$Equal contribution. \textsuperscript{\ding{41}}Corresponding author.}%
\thanks{$^{1}$Robotics Institute, Carnegie Mellon University, Pittsburgh, USA
	{\tt\footnotesize \{uyoo,jmf1,hyaejino,jichnows\}@andrew.cmu.edu}}%
\thanks{$^{2}$University of Michigan, Ann Arbor, USA
{\tt\footnotesize adamhung@umich.edu}}
\thanks{$^{3}$Bosch Center for Artificial Intelligence, Pittsburgh, USA}
}

\begin{document}

\maketitle
\thispagestyle{empty}
\pagestyle{empty}
\renewcommand{\baselinestretch}{0.979} 

%%%%%%%%%%%%%%%%%%%%%%%%%%%%%%%%%%%%%%%%%%%%%%%%%%%%%%%%%%%%%%%%%%%%%%%%%%%%%%%%
\begin{abstract}

 % Humans are capable of continuously manipulating a wide variety of deformable objects into complex shapes, owing largely to our ability to reason about material properties as well as our ability to reason in the presence of geometric occlusion in the object's state; these capabilities allow us to perform diverse tasks such as cooking with dough to expressing ourselves with pottery-making. %However, developing robotic systems that can similarly deform clay robustly remains challenging because of the complex deformation behavior of volumetric deformable objects. 
% Developing robotic systems to robustly perform tasks likfe these remains challenging, however, as they struggle to effectively model and reason about the complex behavior typical of volumetric deformable objects. To study the robotic systems and algorithms capable of deforming volumetric objects, we introduce a novel robotics task of continuously deforming clay on a pottery wheel, and we present a baseline approach for tackling such a task by learning from demonstration. 

Humans are capable of continuously manipulating a wide variety of deformable objects into complex shapes. This is made possible by our intuitive understanding of material properties and mechanics of the object, for reasoning about object states even when visual perception is occluded. These capabilities allow us to perform diverse tasks ranging from cooking with dough to expressing ourselves with pottery-making. However, developing robotic systems to robustly perform similar tasks remains challenging, as current methods struggle to effectively model volumetric deformable objects and reason about the complex behavior they typically exhibit. To study the robotic systems and algorithms capable of deforming volumetric objects, we introduce a novel robotics task of continuously deforming clay on a pottery wheel. We propose a pipeline for perception and pottery skill-learning, called RoPotter, wherein we demonstrate that structural priors specific to the task of pottery-making can be exploited to simplify the pottery skill-learning process. Namely, we can project the cross-section of the clay to a plane to represent the state of the clay, reducing dimensionality. We also demonstrate a mesh-based method of occluded clay state recovery, toward robotic agents capable of continuously deforming clay. Our experiments show that by using the reduced representation with structural priors based on the deformation behaviors of the clay, RoPotter can perform the long-horizon pottery task with 44.4\% lower final shape error compared to the state-of-the-art baselines. Supplemental materials, experiment data, and visualizations are available at \url{robot-pottery.github.io}.

\end{abstract}

%%%%%%%%%%%%%%%%%%%%%%%%%%%%%%%%%%%%%%%%%%%%%%%%%%%%%%%%%%%%%%%%%%%%%%%%%%%%%%%%
%\input{0_outline}
\renewcommand{\baselinestretch}{0.979}

\section{Introduction}

When we perform long-horizon deformable object manipulation tasks such as making pottery, we can continuously manipulate the deformable objects into complex shapes, despite occlusions in the perception of the object's state. This ability to reason about a deformable object's occluded geometry allows us to perform diverse tasks robustly---from cooking with dough to expressing ourselves with pottery. However, developing robotic systems that can robustly perform these tasks remains challenging, due to the complex deformation behavior of volumetric deformable objects~\cite{yin2021modeling}. %In this work, we study robotic systems and algorithms capable of deforming volumetric objects: we introduce a novel robotic task of continuously deforming clay on a pottery wheel.
A common approach to manipulating volumetric deformable objects such as clay has been to learn a dynamics model of the object. 
Despite often remarkable results~\cite{shi2024robocraft,shi2023robocook, bartsch2023sculptbot}, such approaches presently have two drawbacks. First, state-of-the-art methods for learning an explicit dynamics model of the deformable objects in interaction with the environment suffer from a high sample-complexity. For 2-dimensional deformable objects such as cloth, researchers successfully trained the model in simulation and directly transferred it to the real world~\cite{lin2022learning}. However, methods have struggled to similarly model high dimensional interactions due to the increased deformation complexity and the more notable sim-to-real gap in modeling contact dynamics. 
Secondly, and as a result of the first drawback, recent works on volumetric deformable object manipulation often relied on well-parameterized and task-dependent action sequences that allow fully unoccluded observation of the object in between each action to prevent drifting~\cite{shi2024robocraft, bartsch2023sculptbot, shi2023robocook}.

\begin{figure}
    \centering
    \includegraphics[width=1.0\linewidth]{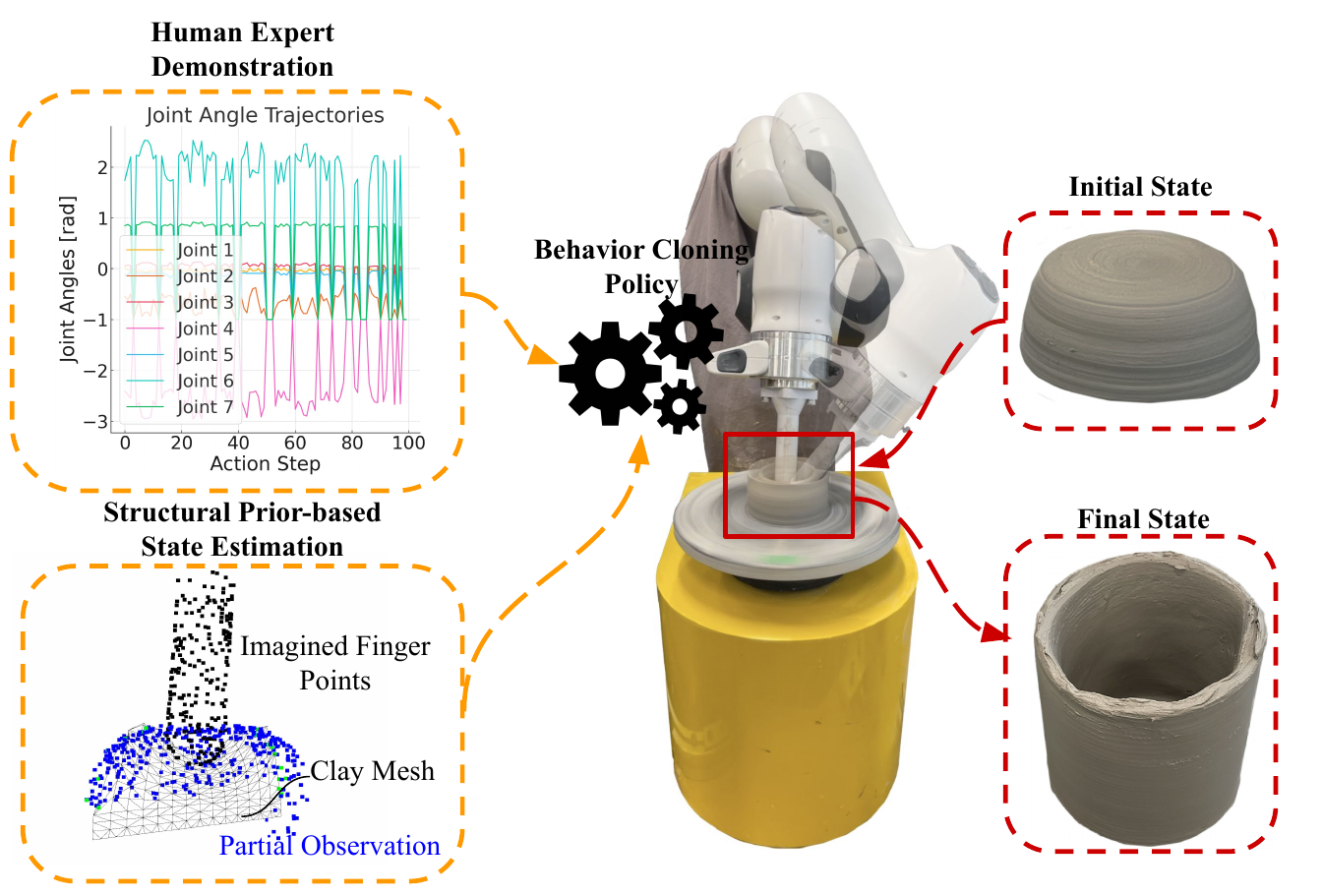}
    \caption{RoPotter Pipeline. We train a behavior cloning policy with expert demonstration data from teleoperation and point cloud state estimates of the clay. We demonstrate that the structural priors of the clay deformations and geometry can assist with the recovery of occluded points during continuous deformable object manipulation.}
    \label{fig:setup}
\end{figure}

Behavior cloning is receiving growing attention among robotic manipulation researchers as it requires fewer expert design choices, such as reward-shaping, as long as the precise actions during demonstrations are known~\cite{torabi2018behavioral, francis2022core, hu2023toward}.
Recently proposed approaches such as Implicit Behavior Cloning~\cite{florence2022implicit} and Diffusion Policy~\cite{chi2023diffusion} have demonstrated an exceptional ability to learn complex manipulation skills given a reasonable number of human demonstrations, with some spatial generalizability and robustness to distractions~\cite{ze20243d}. However, despite their success in diverse manipulation tasks with cloth and clay-rolling---to the best of our knowledge---these works did not generalize these imitation learning pipelines to the manipulation of volumetric deformable objects such as a lump of clay. In this work, we use a pottery wheel with clay to study volumetric deformable object manipulation and demonstrate a pipeline for learning a bowl-making policy with behavior cloning. Our task setup allows researchers to study the space of continuous volumetric deformable object manipulation with a simplified rigid 0 degree-of-freedom end-effector, as shown in Fig.~\ref{fig:setup}. Because clay continuously deforms under contact, staged discrete-action approaches previously taken with volumetric deformable object manipulation tasks ~\cite{bartsch2023sculptbot} are difficult to apply.

\begin{figure}
    \centering
    \includegraphics[width=1.0\linewidth]{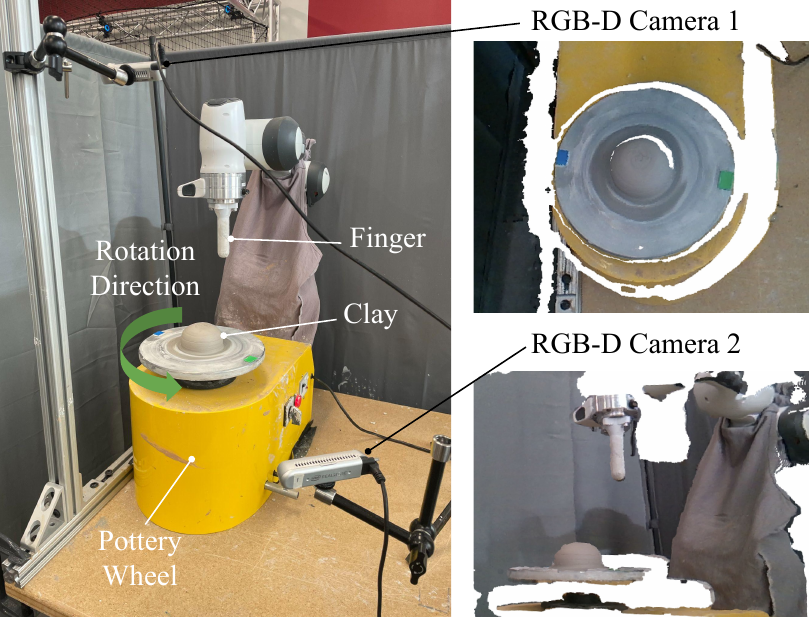}
    \caption{Our setup for robotic pottery-making: two RGB-D cameras each provide a partial view of the clay, and multiple instances of the point clouds are combined to increase how much of the total surface of the clay can be observed.}
    \label{fig:setup}
    \vspace{-0.5cm}
\end{figure}

We present RoPotter, a pipeline for perception and behavior cloning based on diffusion policy~\cite{chi2023diffusion,ze20243d} to deform a block of clay into two bowl shapes. We incorporate structural priors specific to the pottery wheel task to simplify the skill-learning process. Namely, we show that because of the pottery task's radial symmetry, we can reduce the dimension of the explicit clay state representation to the 2D plane by taking a cross-section. Additionally, we demonstrate that we can exploit the structure of a mesh initialized over the starting shape of the clay to reason about the occluded points throughout the task and learn skills effectively. To summarize, we made the following contributions in this work:
\begin{enumerate}
\item We design the novel, long-horizon task of robotic pottery-making wherein an agent deforms a block of clay into two bowls of different dimensions.
\item We develop a system, RoPotter, to collect and register point clouds of the rotating clay deformable object and a pipeline for collecting demonstrations from users.
\item We leverage a structural prior-based approach to shape recovery and estimation of clay undergoing continuous deformation, which we evaluate with ablations.
\item We provide a behavior cloning pipeline for policy learning, which uses the proposed representations of the clay shape, and we evaluate performances with both geometric and semantic metrics.
\end{enumerate}
\renewcommand{\baselinestretch}{0.979} 
\section{Related Work}

\subsection{Deformable Object Perception and Representation}

Conventional methods of representing deformable objects use analytical models such as mass-spring mechanics~\cite{boonvisut2012estimation,guler2015estimating, yin2021modeling}. However, these methods rely on careful system identification, require knowledge about the object's material properties, and are sensitive to empirically-derived parameters~\cite{yin2021modeling, yoo2021analytical}. %
Recent advancements in learning representations with 3D geometry such as PointNet~\cite{qi2017pointnet}, PointNet++~\cite{qi2017pointnet++}, and PointBERT~\cite{yu2022point} have significantly improved robot capabilities in various applications---from autonomous driving to robot manipulation~\cite{bartsch2023sculptbot}. Additionally, methods in the broader family of Graph Neural Networks (GNN) have enabled researchers to introduce structure to the representation and dynamics-learning problem~\cite{shi2024robocraft}. 

Subsequent works demonstrated that the use of priors can introduce structure and thereby regularize the representation learning process, yielding improvements in sample-efficiency of model training~\cite{yoo2024poe}. In the domain of deformable robot shape representation, researchers have noted that mechanics-based priors can help ground the soft body states on physically admissible configurations~\cite{yoo2024poe}. 
However, learning such explicit dynamics models requires either extensive exploration steps to perturb the object and observe state changes~\cite{shi2024robocraft, shi2023robocook}, or exploitation of the simulation environments' privileged information that is not available in the real world~\cite{yoo2023toward}. 
%In this work, 
Instead of explicit dynamics models, we leverage structural prior that does not require laborious steps.
We propose a pipeline for learning deformable object manipulation skills for robot pottery-making, where, inspired by previous works, we incorporate structural priors based on task-dependent mechanics. %; we show that these simplify the problem and address problems of observation occlusion during continuous interactions with deformable objects.

\subsection{Deformable Object Manipulation}
Various works have studied robot manipulation of deformable objects~\cite{yin2021modeling,shi2024robocraft}. Prior works can be broadly categorized by the dimensionality of the deformable objects. Researchers have previously proposed various methods to address robotic manipulation of one-dimensional deformable objects such as ropes and cables~\cite{zhang2021robots,yan_linear_deform}, two-dimensional deformable objects such as cloth and flattened dough~\cite{lin2021softgym}, and three-dimensional deformable objects such as clay and plasticine~\cite{shi2024robocraft}. The ability to reason about manipulating three-dimensional geometries has broad application including in enabling new avenues of human expression through art for instance via human-robot collaborative sculpting~\cite{duenser2020robocut}. In this work, we are specifically interested in robot pottery-making with clay, which falls under the category of three-dimensional deformable object manipulation. To the best of our knowledge, this work presents the first methods for robot pottery-making. 

\begin{figure}
    \centering
    \includegraphics[width=0.8\linewidth]{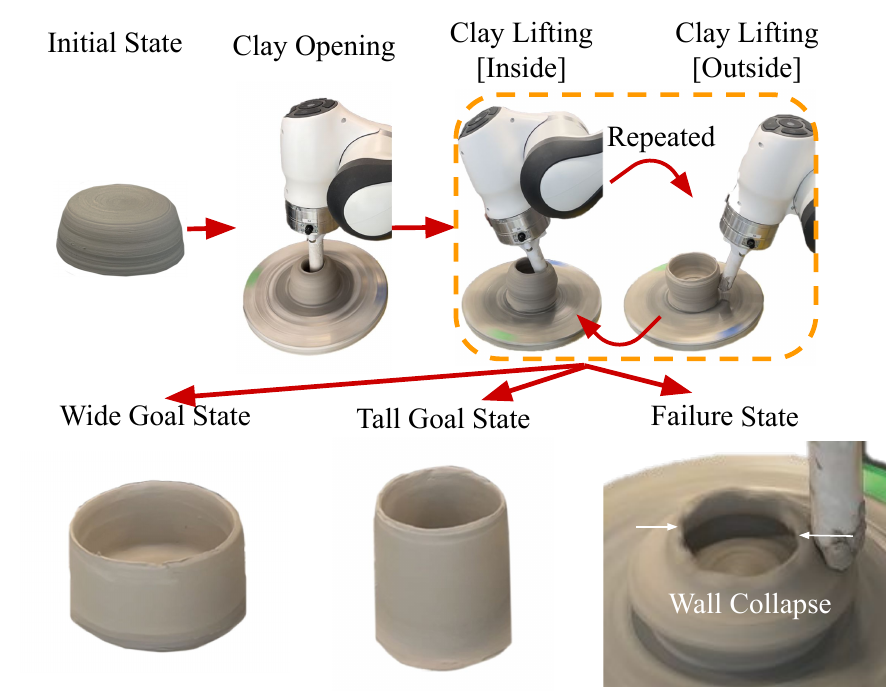}
    \caption{The pottery-making task. We demonstrate the RoPotter's ability to produce two bowls of different dimensions.}
    \label{fig:task}
\end{figure}

\renewcommand{\baselinestretch}{0.979} 
\section{Problem Statement}
\label{sec:problem}

 Let $S$ denote a set of states representing the space of 3D shapes that can be made of clay on a pottery wheel; and $A$, a set of end-effector actions available to a robot. The task of robot pottery-making can be defined as modifying an initial state of clay $s_{i} \in S$ into a desired shape state $s_{g} \in S$, implicitly captured by the demonstrated set of final shape states. The objective here is to learn a policy, $\pi_{s_{g}}(a|s)$, which can prescribe optimal action $a \in A$ defined with respect to the end-effector poses given diverse state $s \in S$ towards achieving goal shape $s_{g} \in S$.

\renewcommand{\baselinestretch}{0.979} 
\section{Methods}
\begin{figure}
    \centering
    \includegraphics[width=0.90\linewidth]{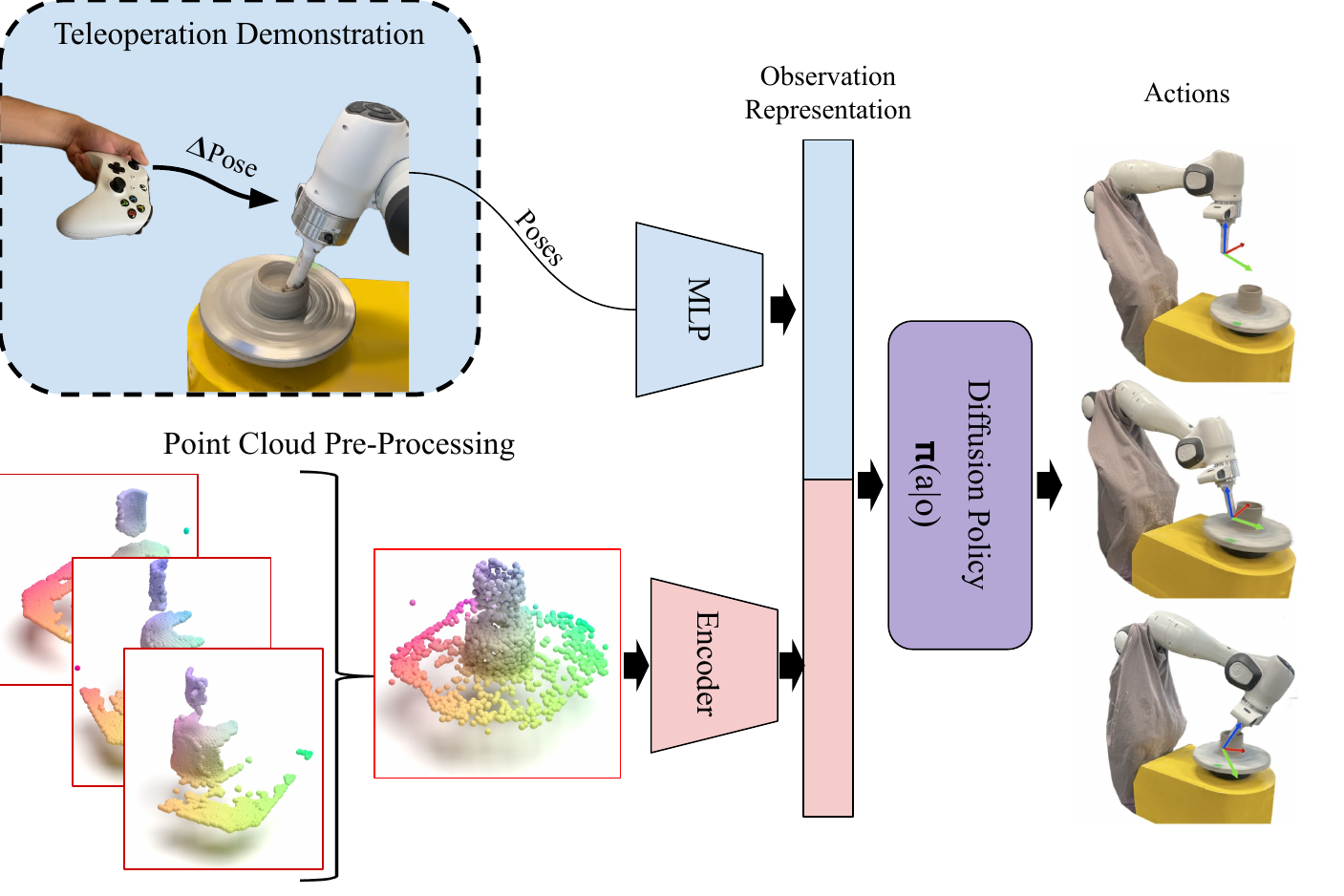}
    \caption{Proposed pipeline for the robotic pottery. We collected expert demonstrations for the two bowl shapes using teleoperation with a gaming console controller. We paired the expert-demonstrated actions with the point cloud observation of the clay shape that was stiched together from multiple virtual perspectives.  }
    \label{fig:pipline}
    \vspace{-0.4cm}
\end{figure}

\begin{algorithm}
 \caption{RoPotter-Mesh Reconstruction}
 \label{alg:ropotter}
 \begin{algorithmic}[1]
 \renewcommand{\algorithmicrequire}{\textbf{Input:}}
 \renewcommand{\algorithmicensure}{\textbf{Output:}}
\Require $P \in \mathbb{R}^{N \times 3}$ \Comment{Initial point cloud}
\Require $x_t \in \mathbb{R}^7$ \Comment{Robot joint angles at time step $t$}
\State $P_{\text{xz}} \gets \{(x, z) \mid (x, y, z) \in P, |y| < T_{\text{thresh}}\}$
\State $H \gets \text{ConvexHull}(P_{\text{xz}})$
\State $H_{\text{aug}} \gets \text{AugmentInteriorPoints}(H)$
\State $M_0 \gets \text{DelaunayTriangulation}(H_{\text{aug}})$ \Comment{Initial mesh}
\State $M_t \gets M_0$
\For{each time step $t$}
    \State $F_t \gets \text{ComputeFingerPoints}(x_t)$
    \State $P_t \gets P \setminus \text{ConvexHull}(F_t)$ \Comment{Remove points in robot finger convex hull}
    \State $P_{\text{xz}, t} \gets \{(x, z) \mid (x, y, z) \in P_t, |y| < T_{\text{thresh}}\}$
    
    \State $M_t \gets \text{MoveVerticesToFit}(M_t, P_{\text{xz}, t})$ \Comment{Move vertices to fit observable points}
    \For{each vertex $v \in M_t$}
        \If{$v \in \text{ConvexHull}(F_t)$}
            \State $n \gets \text{SurfaceNormal}(F_t)$
            \State $v \gets v + \delta \cdot n$ \Comment{Push vertex in the direction of surface normal}
        \EndIf
    \EndFor
    \State $M_t \gets \text{ARAPUpdate}(M_t, M_0)$ \Comment{Update the mesh using ARAP algorithm}
\EndFor

\State \textbf{return} $M_t$
 \end{algorithmic} 
 \end{algorithm}

\begin{table}
\begin{threeparttable}
\caption{RoPotter-Mesh Reconstruction Accuracy Ablation}
\setlength\tabcolsep{0pt} % make LaTeX figure out intercolumn spacing
\begin{tabular*}{\columnwidth}{@{\extracolsep{\fill}} ll ccc}

\toprule
    Method & Input & 
     \multicolumn{2}{c}{CD [mm] $\downarrow^*$} \\ 

     & &  CS & FP \\
\midrule
     % DeepSoRo & Servo Positions & 14.66 & 11.90\\
     Reference CS [Observed] & $P$  & 0.0  & 2.46 \\  \addlinespace\cdashline{1-4} \addlinespace
         Mesh w/o Contact & $P$ &   15.41 & 17.01 \\
\addlinespace
         Mesh w/o ARAP & $x_t, P$ &   4.45 & 7.74 \\
\addlinespace
    Mesh [Proposed] & $x_t, P$  & \bf 1.50 & \bf 2.58\\
\addlinespace

% \addlinespace
%     POE-Graph & Audio [All]&-&  0.00\\
% \addlinespace
%      POE-Graph & Audio [All]&- &  0.00\\
\bottomrule
\end{tabular*}

\scriptsize    
\label{tab:ablation} 
\end{threeparttable}
\begin{minipage}{12cm}
\vspace{0.1cm}
\small $^*$CS refers to cross-section and FP refers to full point cloud. 
\end{minipage}
\end{table}

% Method section intro
We propose a learning from demonstration (LfD) approach to develop a pottery-making robotic system. Our approach is designed to address the sample complexity issue of LfD as well as the occlusions often present during continuous deformable object manipulation. Specifically, structural and mechanics-based priors can help simplify the robot learning problem by grounding predictions to the set of physically permissible spaces~\cite{shi2024robocraft, yoo2024poe}. 
%In RoPotter, we take advantage of the structural priors of the pottery task to reduce the complexity of the problem and account for occlusion.
In the RoPotter approach, we propose the use of two types of structural priors of the pottery task to reduce the complexity of the problem and account for occlusion. 

%To make the learning process more sample-efficient, our approach additionally simplifies the pottery-making task through dimensionality-reduction and leverages structural priors to enable robustness to visual occlusion.

\subsection{RoPotter-2D: Compact Representation with Structural Priors}
\label{sec:prior}
Because of the rotating base of the clay on a pottery wheel, we assume radial symmetry of the clay shapes during the task. We also assume quasi-static conditions for the clay and ignore dynamic effects during manipulation, modeling the clay shapes as discrete states at each time step. 

In our first proposed pipeline, RoPotter-2D, we reduce the dimension of the observation space by taking a cross section composed of the points within a 5mm threshold of a plane containing the clay's center of mass. The reduction of dimension has the benefit of reducing the complexity of the dataset that the policy network must reason over and filtering out uninformative features. The 2D cross-section of the clay concisely captures a corresponding 3D shape state in terms of the diameter, height, and thickness of the walls.
%that largely define the task as outlined in Section~\ref{sec:problem}. 

\subsection{RoPotter-Mesh: Shape Estimation with Structural Priors}
\label{sec:prior-mesh}

In our RoPotter-Mesh reconstruction pipeline as outlined in Algorithm~\ref{alg:ropotter}, we use structural priors on how the clay deforms and an assumption of local smoothness of the clay geometry to recover occluded points. Previous works have shown that deformable soft bodies represented by discrete meshes often follow the constraints of local rigidity, also known as As-Rigid-As-Possible (ARAP) deformation~\cite{sorkine2007rigid, yoo2024poe}. 

ARAP includes a penalty on the rotations of the neighboring edges, producing physically admissible mesh manipulation ~\cite{sorkine2007rigid, levi_smooth_2015, yoo2024poe}. The ARAP energy that we minimize during ARAP deformation is given by: %
\begin{dmath}\footnotesize
\label{eq:ARAP}
     E_{\mathrm{smoothed}}(M, M') = \min_{R_1,..., R_m}\sum_{k=1}^m \Big(\sum_{i,j \in e_k}c_{ijk} \| e_{ij} - R_k e_{ij} \|^2 \\
     + \lambda \hat A \sum_{e_l \in N(e_k)} w_{k l} \| R_k - R_l\|^2\Big)
\end{dmath} %
Here, $M,M'$ define the mesh initialized with the point cloud's convex hull (line 4 Algorithm~\ref{alg:ropotter}) and the deformed mesh respectively, $c_{ijk}$ are the cotan weights~\cite{crane2013dgp}, $\lambda$ is the smoothness regularization weight, $R_1, ..., R_m\in SO(3)$ are the local rotations for each of the edges $e_k \in E$ where $m = |E|$, $\hat A$ is the triangle area and $w_{kl}$ are the scalar weight terms defined by the cotan weights of the dual mesh of $e_{kl}$~\cite{crane2013dgp}. As with previous works on using ARAP for shape recovery and deformation, we use a local-global solver to iteratively minimize the energy defined in Eq.~\ref{eq:ARAP}. During the local optimization step, we compute the locally best-fitting rigid transformation $R_k$ for each of the edge simplices to best map to $M_t$ from $M_0$. In the global step, we update the metric positions of the vertices $i,j \in e_k$ with least squares fitting to make them as consistent with rigid transformation $R_k$ as possible. We repeat the local-global steps until convergence within the energy threshold or maximum iteration.  

As outlined in Algorithm~\ref{alg:ropotter}, we use ARAPUpdate to recover occluded points in the partial point cloud of the clay cross-section. With MoveVerticesToFit, we move the vertices of the mesh to their closest neighbors in the partial point cloud observation (line 10 Algorithm~\ref{alg:ropotter}). We also displace the points in contact with the RoPotter finger, that are identified by checking for points in the convex hull of the finger with respect to the surface normal (line 14 Algorithm~\ref{alg:ropotter}). Using the displaced points and the bottom row of points that are attached to the pottery wheel as constraint points, we minimize the energy in Eq.~\ref{eq:ARAP}.

\begin{figure}
    \centering
    \includegraphics[width=0.90\linewidth]{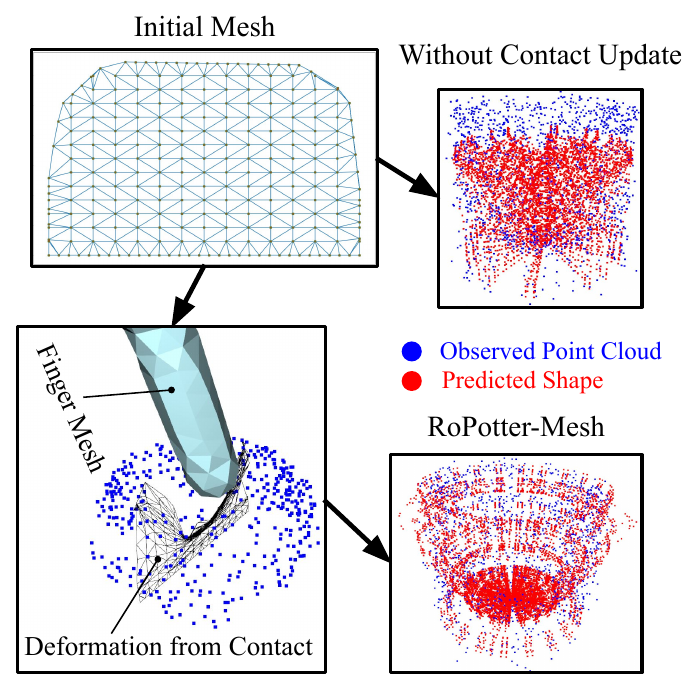}
    \caption{RoPotter-Mesh method outline and ablation results.}
    \label{fig:mesh}
    \vspace{-0.4cm}
\end{figure}
\subsection{Learning From Demonstration with 3D representation}

% MOVED FROM: Related Work: \subsection{Robot Learning from Demonstration}
Learning from Demonstrations via Imitation Learning provides an intuitive approach to robot skill acquisition, where expert demonstrations are directly used by the robot agent policy to quickly learn the involved underlying skills~\cite{schaal1996learning}. In Behavior Cloning, specifically, the policy learns to map observations to the demonstration actions. Such approaches benefit from intuitive training processes, few implementation-level challenges, and strong supervision, at the cost of requiring observations labeled with corresponding expert actions~\cite{torabi2018behavioral}. Our RoPotter approach learns robot pottery-making skills from human demonstration data, using action and observation pairs provided through teleoperation. 

%To make the learning process more sample-efficient, our approach additionally simplifies the pottery-making task through dimensionality-reduction and leverages structural priors to enable robustness to visual occlusion.

\noindent\textbf{Demonstration Collection: }
To collect demonstrations of the robot deforming the clay into a desired bowl shape, we controlled the change in pose of the finger with a gaming console controller (Xbox Wireless Controller, Microsoft) as shown in Fig.~\ref{fig:pipline}. The point clouds are pre-processed and merged into discrete observations, as described in Section~\ref{preprocess}. In each demonstration, we deformed the clay procedurally until the overall measurements of the bowls matched the ones outlined in Section~\ref{sec:problem}. We trained a behavior cloning policy after collecting 40 demonstrations. 

\noindent\textbf{Policy Learning: }
We build on the previous work on diffusion policy with 3D point cloud observations ~\cite{ze20243d} for the presented results. The end-effector orientation is represented with the continuous 6-dimensional representation as previously proposed~\cite{zhou2019continuity}. The full pose of the end-effector with position and orientation is encoded with an MLP to a 64-dimensional feature. The merged and down-sampled point clouds are encoded with the DP3 encoder~\cite{ze20243d} to a 64-dimensional observation feature. For the baseline, we directly use the DP3 encoder architecture as described in Ze, et al.~\cite{ze20243d}. For our two proposed methods, we use the two-dimensional variant of the encoder. Next, the robot pose and point cloud features are concatenated. The diffusion policy then denoises random noise into an action sequence, conditioning on these state features. We use a single clay shape observation as input to the policy and predict a sequence of 8 denoised actions, which helps with the temporal consistency of the trajectories as noted in the literature~\cite{chi2023diffusion, ze20243d}.

\renewcommand{\baselinestretch}{0.979}

\section{Implementation Details} \label{sec:system}
\subsection{Robotic Pottery Setup}
As seen in Fig. \ref{fig:setup}, our robotic pottery setup has a pottery wheel with two colored markers (green and blue) on the edge of the pottery wheel surface that allow us to track the orientation of the clay at a given instance as the clay rotates at approximately three rotations a second. 
%We use a 7 degree-of-freedom robotic arm (Research 3, Franka) with a custom rigid 0 degree-of-freedom finger to manipulate and deform the clay. %We positioned two RGB-D cameras (Realsense D415, Intel), pointed at the clay, to observe the deforming clay shape and to provide approximately top- and side-view perspectives. %We placed a green and blue marker on the edge of the pottery wheel surface to determine the wheel's orientation at a point cloud instance.

\subsection{Sensor Setup}\label{sensors}
Two RGB-D cameras (Realsense D415, Intel) are positioned, pointed at the clay, to observe the deforming clay shape at approximately 25\,Hz and to capture partial point cloud views of top- and side-view perspectives. %\footnote{We reduced the exposure time to minimize the motion-blurring effects at each frame.}. 
The two cameras were calibrated to the pottery wheel frame, where the origin is defined to lie on the wheel's rotation axis. During demonstrations, we capture the robot joint states as well as the point cloud of the bowl from virtual perspectives provided by the rotation of the clay.

\subsection{Point Cloud Pre-Processing}
\label{preprocess}

%If this is the general setup that is exprienced by all models (ours and baselines), then this belongs in the problem statement: The two RGB-D cameras in our setup provide partial point cloud observations of the clay, at approximately 25\,Hz. We reduced the exposure time to minimize the motion-blurring effects at each frame. The two cameras were calibrated to the pottery wheel frame, where the origin is defined to lie on the wheel's rotation axis. 

In the initial frame, we detect both of the markers' centroid positions. We then define the orientation reference vectors from the centroid of each marker to the origin of the pottery wheel frame, and normalize to unit length as $\vec{r_0} = \frac{p_m - p_o}{\Vert p_m - p_o \Vert_2}$ where $p_m, p_o \in \mathbb{R}^2$ are respectively the marker and origin points projected onto the pottery wheel surface. In the subsequent frames, we track one of the two markers' positions to get the new orientation vector $\vec{r_t}$ (the position of the markers is susceptible to occlusion, but both markers are never occluded simultaneously). We then compute the rotation angle $\theta = arccos(\vec{r_t} \cdot \vec{r_0})$. The observed point clouds of the clay are merged as they are rotated from the start orientation until they make a full rotation. The combined point cloud is downsampled to 1,024 points and is used as a single full observation of the clay at a time instance.

\renewcommand{\baselinestretch}{0.979} 

\begin{figure}
    \centering
    \includegraphics[width=1.0\linewidth]{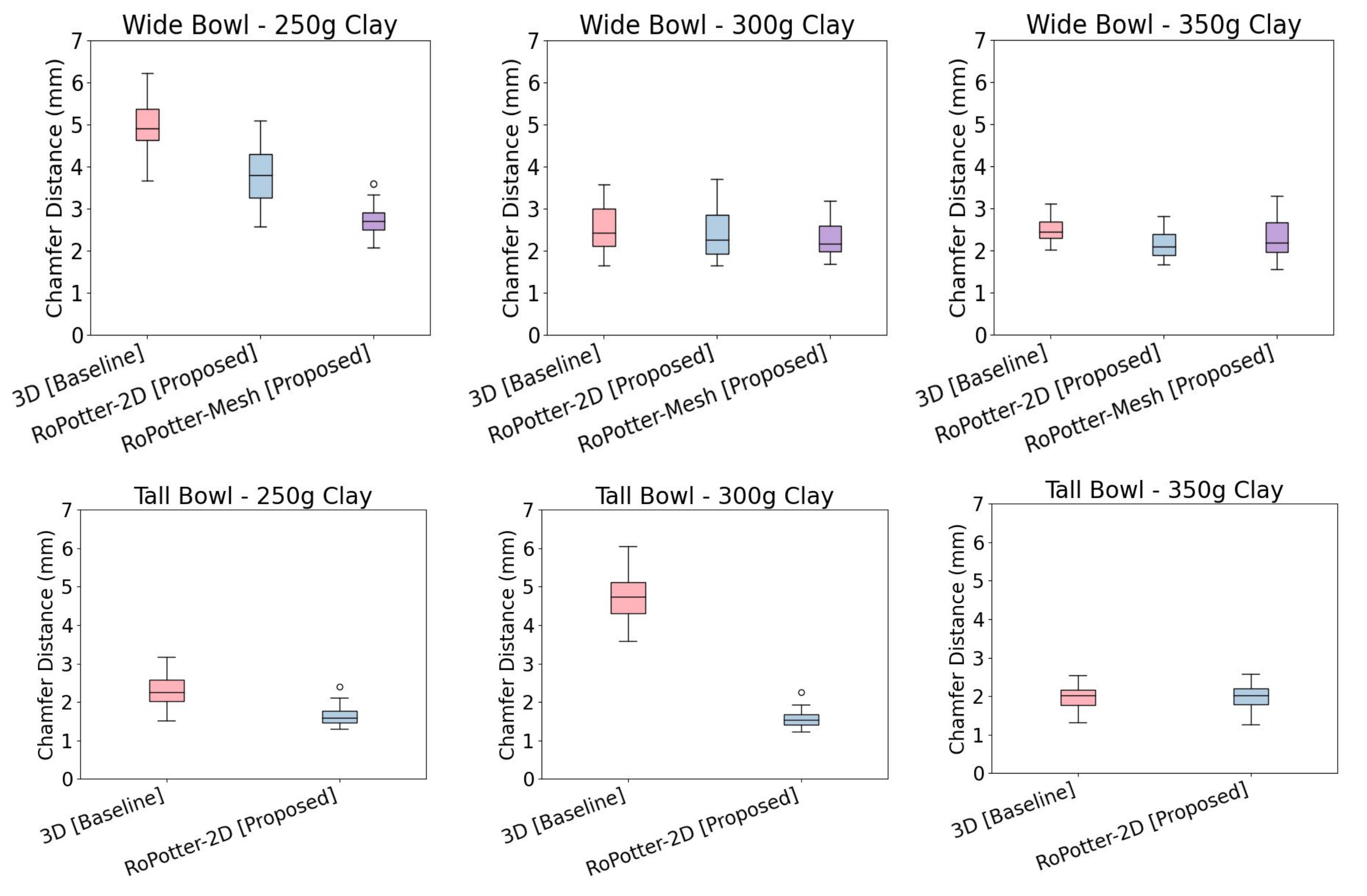}
    \caption{Boxplots showing the full distributions of chamfer distances. For a bowl produced by a given policy and clay mass, the chamfer distance is calculated between this bowl's final state and every final bowl state in the corresponding set of demonstrated bowls, creating a single distribution. }
    \label{fig:boxplots}
    \vspace{-0.4cm}
\end{figure}

\begin{figure*}[t!]
  \centering
     \includegraphics[width=0.9\linewidth]{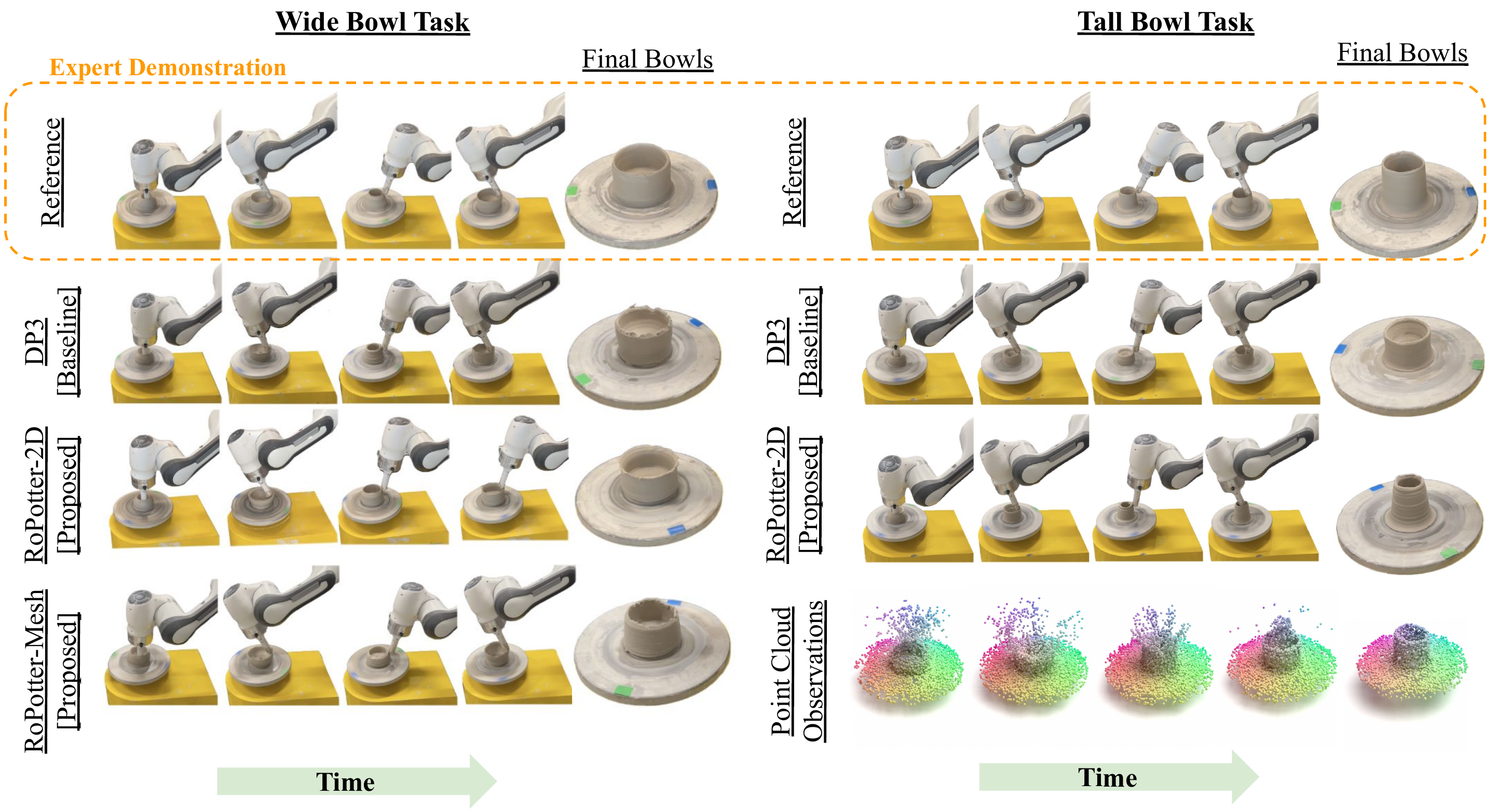}
     \caption{Results of RoPotter robot pottery pipelines. Reference row shows the demonstrated bowls.  }
    \label{fig:results}
\end{figure*}

\begin{table}
\begin{threeparttable}
\caption{Policy Evaluation Against Demonstrated Pottery}
\setlength\tabcolsep{0pt} % make LaTeX figure out intercolumn spacing
\begin{tabular*}{\columnwidth}{@{\extracolsep{\fill}} p{1.9cm} p{1.0cm} p{2.0cm} p{2.2cm} p{1.8cm}} % Adjusted the width of the first column here
\toprule
    Method & Task & Clay Mass [g] & CLIP Similarity $\uparrow$ & CD [mm] $\downarrow$ \\ 
\midrule
\multirow{6}{*}{\makecell{DP3 \\ \textnormal{[Baseline]}}} & \multirow{3}{*}{Wide} & 250 & 0.945 & 4.95 \\ 
    & & 300 & 0.962 & 2.55 \\ 
    & & 350 & 0.971 & 2.50 \\ \cmidrule{2-5}
    & \multirow{3}{*}{Tall} & 250 & 0.965 & 2.32 \\ 
    & & 300 & 0.940 & 4.78 \\ 
    & & 350 & \bf 0.974 & \bf 1.96 \\ 
\midrule
\multirow{6}{*}{\makecell{RoPotter-2D \\ \textnormal{[Proposed]}}} & \multirow{3}{*}{Wide} & 250 & \bf 0.954 & 3.80 \\ 
    & & 300 & \bf 0.972 & 2.44 \\ 
    & & 350 & \bf 0.973 & \bf 2.15 \\ \cmidrule{2-5}
    & \multirow{3}{*}{Tall} & 250 & \bf 0.974 & \bf 1.64 \\ 
    & & 300 & \bf 0.958 & \bf 1.56 \\ 
    & & 350 &  0.968 & \bf 1.99 \\ 
\midrule
\multirow{3}{*}{\makecell{RoPotter-Mesh \\ \textnormal{[Proposed]}}} & \multirow{3}{*}{Wide} & 250 & 0.951 & \bf 2.75 \\ 
    & & 300 & 0.971 & \bf 2.28 \\ 
    & & 350 & 0.972 & 2.28 \\ 
\bottomrule
\end{tabular*}
\scriptsize    
\label{tab:results} 
\end{threeparttable}
\end{table}

\begin{figure}
    \centering
    \includegraphics[width=1.0\linewidth]{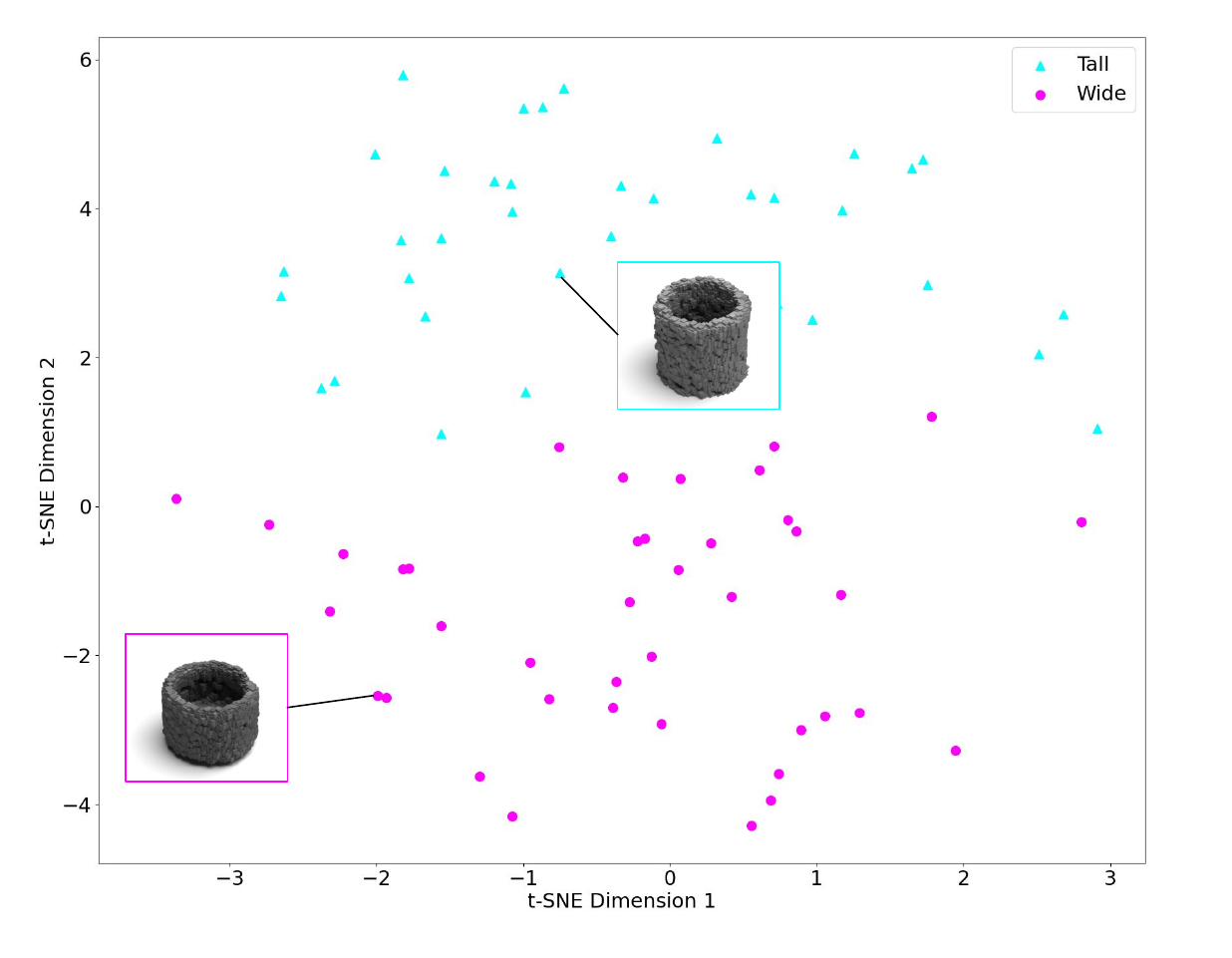}
    \caption{Visualization of the bowl demonstrations' semantic features. }
    \label{fig:clip}
    \vspace{-0.7cm}
\end{figure}

\section{Evaluation}
%For the methods presented in this paper, we evaluated on the two bowl-making tasks as defined in Section~\ref{sec:problem}. 

\subsection{Experimental Setup}\label{sec:exp-setup}

For the purpose of evaluation, we design two types of pottery-making tasks as follows. 
%We introduce the task of robot pottery-making. %We consider a robotic arm with 7 degrees of freedom with a rigid finger end-effector attached to the last link. \jf{This is method/implementation specific}
We define two goal bowl geometries, where a `wide bowl' goal is defined by outer diameter of 100.0\,mm and height of 60.0\,mm, and a `tall bowl' goal has an outer diameter of 70.0\,mm and height of 70.0\,mm. 

All models received 40 demonstrations of teleoperated continuous robot deformable manipulation of clay into each of the two bowl types; demonstrated bowl shapes will have variance caused by imperfect human demonstrations. 

To test the generalizability of the policy, we also controlled for and precisely varied the amount of clay that was on the pottery wheel. We present the metrics that we propose for the evaluation of these policies and discuss the results. 

% \jf{Formulate the generic prediction problem for this task in a way that's agnostic of any specific methodological approach; define the state/action space that all agents will use, describing the inputs and outputs in abstraction. Move the discussion of, e.g., specific hardware/calibration to an `Experimental Setup' subsection (or named paragraph) in the Evaluation section.}

\subsection{Metrics}
\label{sec:metrics}
Previous works on learning from demonstration for manipulation relied on the tasks having clear success criteria, where the researchers can clearly classify the success of trials. However, our novel task of robotic pottery does not afford such clear evaluation criteria. As shown in Fig.~\ref{fig:results}, failure cases can sometimes be clear when the robotic agent fails to continue through the task or causes the catastrophic collapse of the clay structure. In other cases, however, success evaluation may require arbitrary distinction on which mistakes are disqualifying. 
To quantitatively evaluate the proposed methods, we propose two metrics: a geometric similarity and a semantic similarity to the demonstrated pottery. 

\noindent\textbf{Chamfer Distance metric (Geometric Similarity):}
We use the Chamfer Distance (CD) metric for assessing geometric similarity. The unidirectional CD metric from the source point cloud $P_{S}$ to target point cloud $P_{T}$ is defined by~\cite{yoo2023toward, zhang2021unsupervised}:
\begin{equation}
\label{eq:ucd}
    d_{UCD}(P_S, P_T) = \frac{1}{|P_S|}\sum_{x\in P_s}\min_{y\in P_T} \| x-y\|_2
\end{equation}
When elements from both $P_S$ and $P_T$ can be accurately matched to each other, we can use the bidirectional variant of CD defined as the average of $d_{BCD}(P_S, P_T) = \frac{1}{2}[d_{UCD}(P_S, P_T) + d_{UCD}(P_T, P_S)]$.
When we evaluate the RoPotter structural prior-based mesh state estimation, we use $d_{UCD}$ metric from the observed real-world point cloud to the mesh vertex point cloud because the mesh includes internal points of the clay cross-section points that are not visible in the real world with external RGB-D cameras. When we evaluate the real-world point clouds of the bowls produced by the trained policy roll-out against the demonstrated bowl real-world point clouds, we use $d_{BCD}$. 

CD can broadly offer a measurement of geometric accuracy, capturing details such as metric height, diameter, and wall thickness. However, it is also prone to biases from outlier points by averaging the distances in Eq.~\ref{eq:ucd}. CD also tends to filter or average out high-frequency geometric features~\cite{wu2021density}. 

\noindent\textbf{CLIP Score (Semantic Similarity):}
Additionally, in our problem statement in Section~\ref{sec:exp-setup}, goals such as "wide," "tall," and even "bowls" are largely only semantically meaningful descriptors that were implicitly captured with geometric goals of desired diameter and height. Toward evaluating our methods in the semantic space, we propose using CLIP~\cite{radford2021learning} to compute feature similarity between the images of the bowls produced by the policy and the demonstrated bowls. To account for the differing lighting conditions, we re-render the voxelized point clouds using a physics-based renderer~\cite{nimier2019mitsuba} and we use cosine similarity to compare the features. Fig.~\ref{fig:clip} shows a visualization of the feature space spanned by the demonstrated bowls. For both the geometric and semantic metrics, we report the results of the methods in Table~\ref{tab:results}.

\subsection{Experiments}
We first evaluated the RoPotter-Mesh algorithm as outlined in Algorithm~\ref{alg:ropotter} to recover occluded points. For evaluation, we rolled out the demonstrated actions and corresponding partial point clouds for each wide and tall pottery task. We first computed the CD of RoPotter-Mesh's cross-section prediction to the final unoccluded cross-section. We then rotated the predicted cross-section by the pottery wheel's rotational axis and computed the CD to the unoccluded final point cloud of the clay without the finger in the scene. We included ablation results with the contact displacement step and the ARAP step as discussed in Section~\ref{sec:prior-mesh}. To validate that the clay is indeed radially symmetric, we also included a comparison between the unoccluded cross-section that was rotated on the axis of the pottery wheel rotation and compared to the fully observed point cloud of the clay. The observed results in this row provide an upper limit to the achievable performance of the presented methods. 

We compared training a diffusion policy with RoPotter-2D and RoPotter-Mesh against using 3D point clouds as outlined in state-of-the-art 3D diffusion policy work~\cite{ze20243d}. To show the ability of the policy to generalize well to changing initial conditions, we tested the two bowl-making tasks with 250, 300, and 350\,g of clay on the pottery wheel. The initial shape of the clay was consistent across the trials and methods to fairly evaluate the methods.

\subsection{Discussion}

The ablation with the Fig.\ref{fig:mesh} and Table~\ref{tab:ablation} showed that each of the steps in RoPotter-Mesh is necessary to produce reliable mesh recovery of the clay cross-section. We noted that considering the contact conditions of the clay (line 14, Algorithm~\ref{alg:ropotter}) is crucial in reducing the disparity between the reconstructed and ground-truth clay shapes. And introducing the ARAP-based occluded point recovery reduced the CD by 66.3\%.

As shown in the diffusion policy roll-out in Fig.~\ref{fig:results}, the policies trained in this work on 40 demonstrations can effectively deform the lump of clay into a bowl. Notably, the policy learned to replicate the strategy from most of the demonstrations, where we started with pushing the clay out from the center and then iterated on lifting the wall of the bowl from either side until the bowl had the approximately desired diameter and height. 

When the robot finger first opens up the lump of clay into a concave shape, the resulting outer diameter depends on the amount of clay present, because the diameter is governed by the amount of clay available to be displaced laterally. Because the initial outer diameter varies with clay mass, different trajectories are required for the following phase, during which the robot finger lifts the clay walls up from either side. If these trajectories are not adjusted accordingly, the resulting shape may still have the same diameter, as the robot finger can directly constrain this dimension. However, the height may be significantly different. This is because the height of the bowl depends on how much the robot finger pushes on the walls of the bowl.

In our experiments, we observe distinct differences in the policies' abilities to adjust their trajectories given initial clay masses that are slightly out of the distribution of their training data. Across the board, the baseline policy seemed to adjust its trajectory less to the observed initial state, consistently resulting in final bowl shapes that were too tall when a large amount of clay was provided, and too short when a small amount of initial clay was provided. The bowls produced by the 2D and mesh encoder policies also had the same challenges but to a lesser degree. Most noticeably, the baseline policy failed in one trial for a tall bowl with 300g of clay. Quantitatively, we show in Fig.~\ref{fig:boxplots} that for all tasks performed with 250g or 350g of clay, the CD distribution between the policy-produced bowls and demonstration bowls is lower for the bowls produced by the two proposed policies. 

We hypothesized that the policy trained with RoPotter-Mesh would have an advantage when it came to reasoning about the internal shape of the deformed clay as it directly reasons about the contact between the finger and the clay to update the clay states even when it may not be visible from the external perspective. With the experiments with wide bowl-making task, the hypothesis was validated. Additionally, since the mesh is deformed over time but retains information between timesteps, RoPotter-Mesh most likely acted as a filter for observation noise. This may help the continuity of the predicted trajectories over time, and qualitatively we noticed the mesh policy seemed to progress in the tasks more quickly than the other policies with fewer erratic actions compared to the baseline and RoPotter-2D policies. To summarize, the RoPotter-Mesh policy performed 44.4\% better on the wide bowl-making task with 250\,g of clay compared to the baseline 3D policy and overall performed best in the wide bowl-making task. 

The baseline policy only performed best on the tall bowl-making task with 350\, g of clay; however, the proposed methods also performed exceptionally well on this task, achieving CD error of below 2.0\,mm. Notably, we can also observe in Fig.~\ref{fig:boxplots} that the demonstrated bowls that we use to compare the policy-made bowls introduce a range of around 2.0\,mm, indicating that all three policies may have reached upper-limit of performance based on the variance of the demonstrated bowls. The semantic metric scores were consistent with the geometric metric, which can be explained by the fact that the semantic goals of pottery-making were captured well by the geometric task definition.

%Given a similar amount of clay, lack of information about the important geometric features of the clay may not be a problem, as the outer surface of the clay is already a strong signal for the wall thickness when the amount of clay is constrained. However, as we varied the amount of clay on the pottery wheel, we observed that the lack of observation of the internal shape of the deformed clay presented challenges as the bowls produced by the policy began to diverge from the demonstrated bowls.

\renewcommand{\baselinestretch}{0.979} 

\section{Conclusion}
In this work, we present RoPotter, a behavior cloning pipeline for learning a policy for the novel task of robot pottery-making, aided by dimensional reduction of the state space and structural priors. We demonstrated that the RoPotter-Mesh pipeline can recover the occluded points of the clay during continuous clay manipulation tasks with a pottery wheel by relying on mechanics-based priors. The results also showed that using the cross-sectional representation of the clay shape with RoPotter-2D and RoPotter-Mesh allowed the behavior cloning policies trained on them to achieve better performance on the pottery-making tasks even with varying initial conditions.

A limitation of this work is that the RoPotter-Mesh reconstruction method as presented in Algorithm~\ref{alg:ropotter} relies on well-initialized meshes and consistent sampling of accurately observed robot actions to update the cross-section mesh. In this work, these are viable assumptions as the environment is controlled and structured. However, as we extend this pipeline toward a generalizable method for volumetric deformable object manipulation, we plan to integrate the ARAP-based local rigidity into a learned reconstruction approach that will be robust to inconsistent signals.

Another extension of the work will be toward learning goal-conditioned policies for RoPotter. Similar to the challenges of the metrics defined in Section~\ref{sec:metrics}, defining what is an intuitively useful goal representation for the bowl will be a challenge that we will first approach with optimizing over the pre-trained vision-language representations such as CLIP~\cite{radford2021learning}. We also hope to build on these works toward enabling human-robot collaborative sculpting similar to the recent emergence of collaborative human-robot painting~\cite{schaldenbrand2024cofrida}.
\renewcommand{\baselinestretch}{0.979} 
\section*{Acknowledgements}
This work is supported by NSF Graduate Research Fellowship under Grant No. DGE2140739 and by the Technology Innovation Program (20018112, Development of autonomous manipulation and gripping technology using imitation learning based on visual tactile sensing) funded by the Ministry of Trade, Industry and Energy (MOTIE, Korea). We thank John Zhang, Sofia Kwok, Eliot Xing, Jeong Hun Lee and Peter Schaldenbrand for the discussions and feedback.

%%%%%%%%%%%%%%%%%%%%%%%%%%%%%%%%%%%%%%%%%%%%%%%%%%%%%%%%%%%%%%%%%%%%%%%%%%%%%%%%

\addtolength{\textheight}{-0.2cm}   % This command serves to balance the column lengths

%%%%%%%%%%%%%%%%%%%%%%%%%%%%%%%%%%%%%%%%%%%%%%%%%%%%%%%%%%%%%%%%%%%%%%%%%%%%%%%%

%\clearpage

{\footnotesize
\bibliographystyle{ieeetr}
\bibliography{references.bib}
}

\end{document}